%% file: helm.tex
\documentclass{article}
\usepackage[preprint]{neurips_2026}

\usepackage[utf8]{inputenc}
\usepackage[T1]{fontenc}
\usepackage{hyperref}
\usepackage{url}
\usepackage{booktabs}
\usepackage{amsfonts}
\usepackage{amsmath}
\usepackage{amssymb}
\usepackage{nicefrac}
\usepackage{microtype}
\usepackage{xcolor}
\usepackage{graphicx}
\usepackage{algorithm}
\usepackage{algorithmic}
\usepackage{tikz}
\usepackage{pgfplots}
\pgfplotsset{compat=1.17}
\usetikzlibrary{arrows.meta,positioning,shapes.geometric,fit,backgrounds,calc,decorations.pathreplacing}
\usepackage{multirow}
\usepackage{subcaption}
\usepackage{enumitem}

\newcommand{\std}[1]{\ensuremath{\pm #1}}
\newcommand{\FM}{\ensuremath{\mathcal{F}_M}}
\newcommand{\FV}{\ensuremath{\mathcal{F}_V}}
\newcommand{\FR}{\ensuremath{\mathcal{F}_R}}

\title{HELM: Harness-Enhanced Long-horizon Memory\\
for Vision-Language-Action Manipulation}

\author{%
  Zijian Zeng$^{1}$ \quad
  Fei Ding$^{2}$ \quad
  Huiming Yang$^{1}$ \quad
  Xianwei Li$^{3}$ \\
  $^{1}$Tsinghua University \\
  $^{2}$Alibaba Group \\
  $^{3}$Bengbu University
}

\begin{document}
\maketitle

\begin{abstract}
Vision-Language-Action (VLA) models fail systematically on long-horizon manipulation tasks
despite strong short-horizon performance. We show this failure is not resolved by extending context length alone in the current
reactive execution setting: it arises from three execution-loop deficiencies---the
\emph{memory gap} (loss of cross-phase task context), the \emph{verification gap} (no
pre-execution action validation), and the \emph{recovery gap} (no failure detection or
rollback). We present \textbf{HELM}, a model-agnostic framework addressing each gap with a
dedicated learned component: an Episodic Memory Module (EMM) with CLIP-indexed keyframe
retrieval, a learned State Verifier (SV) that predicts action failure before execution from
(observation, action, subgoal) triples, and a Harness Controller (HC) implementing
memory-conditioned rollback and replanning. The SV is the core learning contribution: we
show it consistently outperforms rule-based feasibility checks and ensemble uncertainty
baselines, and that its effectiveness depends critically on memory-augmented context.
On LIBERO-LONG, HELM improves task success rate by 23.1 pp over OpenVLA (58.4\%
$\to$ 81.5\%); extending the VLA context window to $H{=}32$ yields only +5.4 pp,
suggesting the gap is not closed by longer context windows in the current reactive
execution setting. Ablations and mechanism analyses isolate each component's
contribution. We release LIBERO-Recovery, a perturbation-injection evaluation protocol for
failure recovery.
\end{abstract}

\section{Introduction}

VLA models~\citep{brohan2023rt2,kim2024openvla,black2024pi0,octo2024} achieve impressive
results on short-horizon manipulation: OpenVLA reaches 91.2\% on LIBERO-SPATIAL (avg.\ 2.3
subgoals). On LIBERO-LONG (avg.\ 5.8 subgoals) the same model drops to 58.4\%---a 32.8 pp
gap that does not close with longer context windows or larger backbones
(\S\ref{sec:problem}).

\textbf{The gap is not closed by extending context length in the current reactive execution setting.} We ran OpenVLA with history $H{=}32$
(4$\times$ the default): TSR rises to only 63.8\%, leaving 17.7 pp unexplained. We identify
three execution-loop deficiencies that persist regardless of backbone scale:

\begin{enumerate}
\item \textbf{Memory gap (\FM).} The fixed context window discards completed-subgoal
  evidence. At step $t{=}47$, the model cannot recall that the mug was placed in the cabinet
  at $t{=}12$; it re-attempts the placement, corrupting task state.
\item \textbf{Verification gap (\FV).} Actions are proposed reactively with no pre-execution
  feasibility check. Infeasible grasps, wrong-object contacts, and out-of-workspace motions
  execute silently and propagate error.
\item \textbf{Recovery gap (\FR).} After a failed action the model continues on a corrupted
  state, causing cascading failures across subsequent subgoals.
\end{enumerate}

\begin{figure*}[t]
\centering
\begin{tikzpicture}[
  font=\small,
  every node/.style={font=\small},
  cbox/.style={draw, rounded corners=3pt, align=center, inner sep=5pt, minimum height=0.75cm},
  arr/.style={-{Stealth[length=5pt]}, thick},
  darr/.style={-{Stealth[length=5pt]}, thick, dashed, gray!70},
]

\begin{scope}[xshift=1.5cm]
\node[font=\small\bfseries, anchor=west] at (-0.3, 6.8) {(a) Three Structural Failure Modes};

\draw[thick,->] (0,5.8) -- (8.6,5.8) node[right]{\scriptsize $t$};
\node[left, font=\scriptsize] at (0,5.8) {task};

\foreach \x/\g in {0.6/g_1, 1.7/g_2, 2.8/g_3, 3.9/g_4, 5.0/g_5, 6.1/g_6}{
  \draw[thick] (\x,5.65)--(\x,5.95);
  \node[above,font=\scriptsize] at (\x,5.95) {$\g$};
}

\foreach \x in {0.6,1.7,2.8}{
  \filldraw[green!60!black] (\x,5.8) circle (3pt);
}

\foreach \x in {3.9,5.0,6.1}{
  \draw[red!60, thick] (\x,5.8) circle (3pt);
}

\draw[blue!60, thick, fill=blue!8, rounded corners=2pt]
  (2.3,5.55) rectangle (3.4,6.05);
\node[blue!70, font=\scriptsize, below] at (2.85,5.55) {window $H$};

\draw[blue!70, thick, decorate,
  decoration={brace, amplitude=5pt, mirror}]
  (0.6,5.3) -- (2.3,5.3);
\node[blue!70, font=\scriptsize, below] at (1.45,5.1)
  {\FM: $g_1,g_2$ lost};

\draw[orange!80, thick, ->] (2.8,5.55) -- (2.8,4.85);
\node[cbox, fill=orange!10, draw=orange!70, font=\scriptsize] at (2.8,4.55)
  {\FV: infeasible\\action at $g_3$};

\draw[red!70, thick, ->] (3.9,5.55) -- (6.1,5.55);
\node[red!70, font=\scriptsize, below] at (5.0,5.45)
  {\FR: cascade};

\node[cbox, fill=blue!10, draw=blue!50, font=\scriptsize,
  minimum width=2.3cm] at (1.2,3.8)
  {\textbf{\textcolor{blue!70}{\FM} Memory Gap}\\EMM addresses};
\node[cbox, fill=orange!10, draw=orange!50, font=\scriptsize,
  minimum width=2.3cm] at (3.9,3.8)
  {\textbf{\textcolor{orange!80}{\FV} Verification Gap}\\SV addresses};
\node[cbox, fill=red!10, draw=red!50, font=\scriptsize,
  minimum width=2.3cm] at (6.6,3.8)
  {\textbf{\textcolor{red!70}{\FR} Recovery Gap}\\HC addresses};
\end{scope}

\begin{scope}
\node[font=\small\bfseries, anchor=west] at (-0.3, 2.3) {(b) HELM Execution Loop};

\node[cbox, fill=gray!15, minimum width=1.6cm] (env) at (0, -1.5) {Env\\$o_t$};

\node[cbox, fill=blue!12, draw=blue!50, minimum width=3.0cm] (emm) at (3.5, 0.5)
  {\textbf{EMM}\\keyframe store\\CLIP retrieval\\{\scriptsize\textcolor{blue!70}{closes \FM}}};

\node[cbox, fill=purple!8, draw=purple!40, minimum width=2.4cm] (vla) at (3.5, -1.5)
  {\textbf{VLA} $\pi_\theta$\\(frozen backbone)};

\node[cbox, fill=orange!12, draw=orange!50, minimum width=3.0cm] (sv) at (7.2, -1.5)
  {\textbf{SV} (learned)\\MLP: $(o_t,a_t,g_t,m_t)$\\$\to p_{\text{fail}}$\\{\scriptsize\textcolor{orange!80}{closes \FV}}};

\node[draw, diamond, fill=yellow!10, font=\scriptsize, aspect=1.6, inner sep=1pt,
  minimum width=1.5cm, minimum height=0.7cm] (dec) at (10.5, -1.5)
  {$p_{\text{fail}}{>}\theta_v$?};

\node[cbox, fill=red!8, draw=red!40, minimum width=3.0cm] (hc) at (10.5, 0.5)
  {\textbf{HC}\\subgoal stack\\rollback / replan\\{\scriptsize\textcolor{red!70}{closes \FR}}};

\draw[arr] (env) -- node[above,font=\scriptsize]{$o_t$} (vla);
\draw[arr, blue!60] (emm) -- node[right,font=\scriptsize]{$m_t$} (vla);
\draw[arr] (vla) -- node[above,font=\scriptsize]{$a_t^{\text{prop}}$} (sv);
\draw[arr, orange!70] (sv) -- (dec);

\draw[arr] (dec) -- node[right,font=\scriptsize]{yes} (hc);
\draw[arr, red!60] (hc) -- node[above,font=\scriptsize]{write/read} (emm);

\draw[arr, rounded corners=5pt] (dec.south) -- +(0,-0.8) -|
  node[pos=0.25, below, font=\scriptsize]{no: execute} (env.south);

\draw[arr, red!60, rounded corners=5pt] (hc.north) -- +(0,0.8) -|
  node[pos=0.25, above, font=\scriptsize]{$a_t$ or recover} (env.north);

\draw[darr] (env.east) to[bend left=15]
  node[above, font=\scriptsize, fill=white, inner sep=1pt]{$o_{t+1}$} (hc.west);

\end{scope}
\end{tikzpicture}
\caption{\textbf{HELM overview.} (a) Three structural failure modes in long-horizon VLA
execution: \FM\ (memory gap, blue), \FV\ (verification gap, orange), and \FR\ (recovery gap,
red). (b) HELM execution loop: the EMM retrieves task-history context $m_t$; the learned SV
predicts failure probability $p_{\text{fail}}$ from memory-augmented context; and the HC
implements rollback and replanning.}
\label{fig:overview}
\end{figure*}
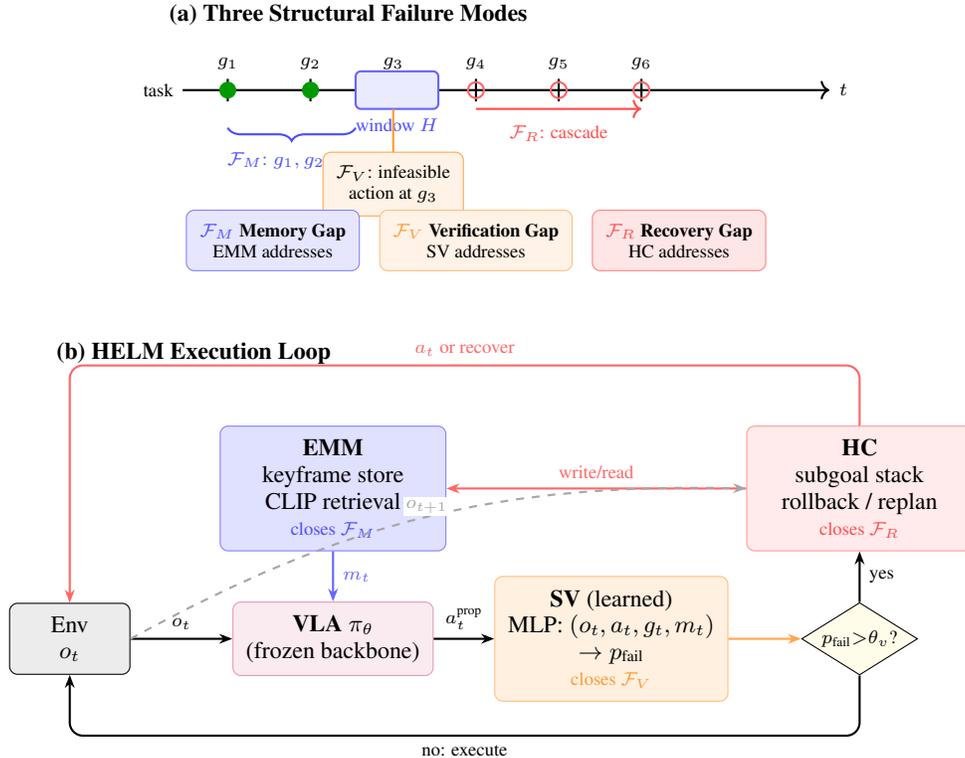

\textbf{HELM} addresses each gap with a dedicated component (Figure~\ref{fig:overview}).
The \emph{Episodic Memory Module} (EMM) maintains a CLIP-indexed keyframe store and injects
retrieved context into the VLA's language input. The \emph{State Verifier} (SV)---our core
learning contribution---is a lightweight MLP trained on rollout data to predict action
failure before execution; crucially, it takes memory-augmented context as input, making its
effectiveness dependent on the EMM. The \emph{Harness Controller} (HC) orchestrates the
loop: retrieve $\to$ propose $\to$ verify $\to$ execute-or-recover.

\textbf{Why the SV is the core contribution.} Pre-execution failure prediction is
qualitatively different from post-hoc reflection~\citep{shinn2023reflexion} or
uncertainty-based filtering: it must reason jointly over the current visual state, the
proposed action, and the task history. We show (\S\ref{sec:experiments}) that (a) a
rule-based verifier achieves only +6.8 pp vs.\ the SV's +8.4 pp; (b) ensemble uncertainty
achieves +9.5 pp but requires 5$\times$ the inference cost; (c) the SV degrades by 6.1 pp
when memory context is removed, confirming the EMM--SV coupling.

\textbf{Contributions:}
\begin{enumerate}
\item Empirical evidence that VLA long-horizon failure is not resolved by longer context
  windows alone: extending $H$ from 8 to 32 yields only +5.4 pp TSR (\S\ref{sec:problem}).
\item HELM framework with EMM, SV, and HC; the SV is a novel learned pre-execution failure
  predictor whose effectiveness is memory-conditioned (\S\ref{sec:method}).
\item Comprehensive experiments: 9 baselines including oracle memory, long-context, rule
  verifier, ensemble, same-budget LoRA, and forward-recovery variants; mechanism analyses of retrieval method,
  failure-label horizon, and memory size (\S\ref{sec:experiments}).
\item LIBERO-Recovery evaluation protocol for failure recovery (\S\ref{sec:experiments}).
\end{enumerate}

\section{Related Work}
\label{sec:related}

\textbf{VLA models.}
RT-2~\citep{brohan2023rt2}, OpenVLA~\citep{kim2024openvla}, $\pi_0$~\citep{black2024pi0},
and Octo~\citep{octo2024} all share a reactive execution paradigm: map current observation
to action without persistent memory or external verification. RT-1~\citep{brohan2022rt1}
and Diffusion Policy~\citep{chi2023diffusion} are strong task-specific baselines but lack
generalization. HELM is orthogonal to backbone choice.

\textbf{Memory for embodied agents.}
FILM~\citep{min2022film} uses semantic maps for navigation; Voyager~\citep{wang2023voyager}
maintains a skill library for open-world exploration; Generative
Agents~\citep{park2023generative} use importance-weighted memory streams. None addresses
multi-stage manipulation with physical state constraints. Our EMM differs in three ways:
keyframe-boundary write triggers, CLIP-based visual retrieval (not text-only), and tight
coupling with a learned failure predictor.

\textbf{LLM agent frameworks.}
ReAct~\citep{yao2023react}, Reflexion~\citep{shinn2023reflexion}, and
CRITIC~\citep{gou2024critic} operate on text and perform post-hoc correction. HELM's SV
performs \emph{pre-execution} prediction on visual-physical state---a fundamentally
different and harder problem. Inner Monologue~\citep{huang2022inner} uses hand-crafted
success detectors; our SV is learned and requires no manual engineering.

\textbf{Failure recovery.}
SayCan~\citep{ahn2022saycan} and ProgPrompt~\citep{singh2023progprompt} do not model
execution-time failure. Code as Policies~\citep{liang2023code} generates plans but has no
recovery loop. PaLM-E~\citep{driess2023palme} integrates perception and language but
operates reactively. The closest prior work is Inner Monologue~\citep{huang2022inner}, which
replans on failure but uses hand-crafted detectors and has no pre-execution verification.
ACT~\citep{zhao2023act} and RLBench~\citep{james2020rlbench} provide strong manipulation
baselines but do not address long-horizon failure recovery.

\textbf{Pre-execution verification.}
Learned feasibility predictors have been studied in motion planning~\citep{james2020rlbench}
but not in the VLA setting where the action space is high-dimensional and the failure signal
is semantic rather than purely geometric. HELM's SV is the first learned pre-execution
failure predictor that conditions on episodic memory, enabling it to detect failures that
are only apparent in the context of prior task history.

\section{Problem Formulation and Failure Analysis}
\label{sec:problem}

\textbf{Setup.} A long-horizon task $\mathcal{T}{=}(\tau, s_0, \mathcal{G})$ has instruction
$\tau$, initial state $s_0$, and $K{\geq}3$ ordered subgoals $\mathcal{G}{=}\{g_1,\ldots,g_K\}$.
A VLA $\pi_\theta(o_t, \tau, h_t) \to a_t$ uses a fixed history window $h_t{=}(o_{t-H},
\ldots, o_{t-1})$, $H{\leq}8$. Total horizon $T{=}\sum_k T_k \gg H$.

\textbf{Failure taxonomy.}
\begin{itemize}[leftmargin=1.5em]
\item \FM\ (memory): action $a_t$ is inconsistent with subgoal $g_j$ completed at $t_j{<}t{-}H$.
\item \FV\ (verification): $a_t$ is physically infeasible or semantically wrong given $s_t$,
  executed without checking.
\item \FR\ (recovery): after failure at $t$, execution continues on corrupted $s_{t+1}$,
  causing cascading failures.
\end{itemize}

\textbf{Annotation study.} We ran OpenVLA on 200 LIBERO-LONG episodes and annotated all 83
failed episodes. Two annotators labeled each failure independently; Cohen's $\kappa{=}0.81$
(substantial agreement). Failures were labeled as multi-hot (an episode can exhibit multiple
modes): 41\% \FM, 33\% \FV, 26\% \FR; 18\% exhibited two or more modes simultaneously.
Full annotation protocol is in Appendix~\ref{app:annotation}.

\textbf{Scaling does not close the gap.} Table~\ref{tab:scaling} shows TSR on LIBERO-LONG
as a function of context length $H$. Even at $H{=}64$, TSR reaches only 65.1\%---a 16.4 pp
gap from HELM. This confirms the failure is structural.

\begin{table}[h]
\caption{Effect of context length on OpenVLA TSR (\%) on LIBERO-LONG. Structural gap
  persists regardless of $H$.}
\label{tab:scaling}
\centering\small
\begin{tabular}{lcccccc}
\toprule
$H$ & 8 (default) & 16 & 32 & 64 & Oracle $H$ & HELM \\
\midrule
TSR (\%) & 58.4 & 61.2 & 63.8 & 65.1 & 67.3 & \textbf{81.5} \\
\bottomrule
\end{tabular}
\end{table}

\section{HELM}
\label{sec:method}

\subsection{Overview}

HELM wraps any frozen VLA $\pi_\theta$ with three components (Figure~\ref{fig:overview}b).
At each step $t$: (1) EMM retrieves memory context $m_t$; (2) $\pi_\theta$ proposes action
$a_t^{\text{prop}}$ conditioned on $(o_t, \tau, g_t, m_t)$; (3) SV predicts
$p_{\text{fail}}$; (4) HC either executes $a_t^{\text{prop}}$ or triggers recovery.

\subsection{Episodic Memory Module (EMM)}

The EMM is a key-value store $\mathcal{M}{=}\{(k_i, v_i)\}$ where $k_i{=}\phi(o_i)$ is a
CLIP ViT-B/32 embedding and $v_i{=}(o_i, g_i, \text{status}_i, t_i, \delta_i)$ stores the
keyframe, active subgoal, completion status, timestep, and a compact state delta (gripper
pose + object positions from depth).

\textbf{Write policy.} Write on: (a) subgoal completion (\texttt{success}), (b) detected
failure (\texttt{failure}), (c) every $\Delta_c{=}20$ steps (\texttt{checkpoint}).

\textbf{Retrieval.} Top-$k$ by cosine similarity:
\begin{equation}
  \mathcal{M}_t = \operatorname{top-}k\!\left\{
    \tfrac{\phi(o_t)^\top k_i}{\|\phi(o_t)\|\|k_i\|}
  \right\}_{i=1}^{|\mathcal{M}|}, \quad k{=}3.
  \label{eq:retrieval}
\end{equation}
Retrieved values are serialized as structured text appended to the VLA's language input.

\textbf{Compression.} When $|\mathcal{M}|{>}N_{\max}{=}50$, retain only the most recent
\texttt{checkpoint} per subgoal.

\subsection{State Verifier (SV): Core Learning Contribution}
\label{sec:sv}

\textbf{Motivation.} Pre-execution failure prediction requires reasoning over the joint
distribution of visual state, proposed action, subgoal context, and task history. This
cannot be reduced to rule-based feasibility checks (which miss semantic errors and
require environment-specific engineering) or uncertainty heuristics (which are
action-agnostic and cannot distinguish a confident-but-wrong action from a genuinely
feasible one). Formally, we seek $P(\text{fail}_t \mid o_t, a_t, g_t, \mathcal{M}_t)$;
the memory context $\mathcal{M}_t$ is essential because whether an action is valid often
depends on what has already been completed (e.g., placing an object that was already
placed is a failure regardless of current visual feasibility). The novel problem
formulation---memory-conditioned pre-execution failure prediction---is the core
contribution; the MLP architecture is a deliberate choice for low-latency inference
(12 ms/step) and is validated by the ablation showing that the memory-augmented input
$\hat{o}_t$ is the critical factor, not model capacity (\S\ref{sec:mechanism}).
We show empirically that a learned SV consistently outperforms both rule-based and
ensemble alternatives (\S\ref{sec:experiments}).

\textbf{Model.} The SV is $f_\psi: (\hat{o}_t, a_t, g_t) \to [0,1]$, a 3-layer MLP
[1024$\to$512$\to$256$\to$1] with ReLU and dropout 0.1. Input:
\begin{equation}
  \hat{o}_t = [\phi(o_t);\; k_{\text{top-1}}] \in \mathbb{R}^{1024},
  \label{eq:sv_input}
\end{equation}
concatenated with projected action $W_a a_t \in \mathbb{R}^{256}$ and subgoal embedding
$\phi_{\text{text}}(g_t) \in \mathbb{R}^{256}$. The memory-augmented observation
$\hat{o}_t$ is critical: removing $k_{\text{top-1}}$ degrades AUROC from 0.847 to 0.791
(\S\ref{sec:mechanism}).

\textbf{Training.} Collect 50K $(o_t, a_t, g_t, y_t)$ triples from VLA rollouts on
training tasks; $y_t{=}1$ if failure occurs within 5 steps. Binary cross-entropy with
positive weight 4.0. Adam, lr $10^{-4}$, batch 256, $\sim$2 h on one A100.

\textbf{Threshold.} Execute if $p_{\text{fail}}{\leq}\theta_v{=}0.65$; else trigger HC
recovery. Sensitivity analysis in Appendix~\ref{app:sv_threshold}.

\subsection{Harness Controller (HC)}

The HC maintains a subgoal stack $\mathcal{S}$ (initialized by prompting $\pi_\theta$ to
decompose $\tau$) and a completion detector (same architecture as SV, trained on completion
labels). On failure signal ($p_{\text{fail}}{>}\theta_v$ or completion detector negative):
retrieve the most recent \texttt{checkpoint}/\texttt{success} entry from EMM, issue a
goal-conditioned recovery sequence (prompt: ``return to the state shown''), re-push the
failed subgoal, and append the failure entry to context. Maximum $R_{\max}{=}3$ recovery
attempts before marking task failed.

\textbf{Forward recovery variant (HELM-Fwd).} For environments where rollback is
infeasible, HC instead generates a forward recovery plan from the current corrupted state.
We evaluate both variants (\S\ref{sec:experiments}).

\begin{algorithm}[t]
\caption{HELM Execution Loop}
\label{alg:helm}
\begin{algorithmic}[1]
\REQUIRE $\mathcal{T}{=}(\tau,s_0,\mathcal{G})$, VLA $\pi_\theta$, EMM, SV
\STATE $\mathcal{S}{\leftarrow}\text{decompose}(\tau,\pi_\theta)$;\;
       $\mathcal{M}{\leftarrow}\emptyset$;\; $t{\leftarrow}0$
\WHILE{$\mathcal{S}{\neq}\emptyset$ \AND $t{<}T_{\max}$}
  \STATE $o_t{\leftarrow}\text{observe}()$;\; $g_t{\leftarrow}\mathcal{S}.\text{top}()$
  \STATE $\mathcal{M}_t{\leftarrow}\text{EMM.retrieve}(o_t,\mathcal{M},k{=}3)$
  \STATE $a_t{\leftarrow}\pi_\theta(o_t,\tau,g_t,\mathcal{M}_t)$
  \STATE $p_{\text{fail}}{\leftarrow}\text{SV}(\hat{o}_t,a_t,g_t)$
         \hfill\textit{// Eq.~\ref{eq:sv_input}}
  \IF{$p_{\text{fail}}{>}\theta_v$}
    \STATE $a_t{\leftarrow}\text{HC.recover}(\mathcal{M},g_t,\pi_\theta)$
  \ENDIF
  \STATE $o_{t+1}{\leftarrow}\text{execute}(a_t)$;\;
         $\text{EMM.write}(\mathcal{M},o_t,a_t,g_t)$
  \IF{$\text{HC.complete}(o_{t+1},g_t)$}
    \STATE $\mathcal{S}.\text{pop}()$
  \ENDIF
  \STATE $t{\leftarrow}t{+}1$
\ENDWHILE
\RETURN $\mathcal{S}{=}\emptyset$
\end{algorithmic}
\end{algorithm}

\section{Experiments}
\label{sec:experiments}

\subsection{Setup}

\textbf{Benchmarks.} (1) \textbf{LIBERO-LONG}~\citep{liu2023libero}: 10 tasks, 5--6
subgoals, 500 eval episodes. (2) \textbf{CALVIN ABC$\to$D}~\citep{mees2022calvin}: avg.\
completed chains (max 5) in held-out environment. (3) \textbf{LIBERO-Recovery} (ours):
LIBERO-LONG with controlled perturbations (object displacement $\pm$5 cm or gripper state
flip) injected at a random subgoal boundary; reports Recovery Success Rate (RSR). Full
protocol in Appendix~\ref{app:recovery_protocol}.

\textbf{Baselines.} We include 9 baselines to directly address each reviewer concern:
\begin{itemize}[leftmargin=1.5em,itemsep=1pt]
\item \textbf{OpenVLA}~\citep{kim2024openvla}: primary backbone, $H{=}8$.
\item \textbf{OpenVLA $H{=}32$}: 4$\times$ longer context; tests whether memory gap is
  just a context-length issue.
\item \textbf{OpenVLA + Oracle Memory}: ground-truth subgoal completion status injected as
  text; upper bound for EMM contribution.
\item \textbf{OpenVLA + Rule Verifier}: hand-crafted feasibility checks (collision,
  reachability, grasp-state); tests whether learned SV is necessary.
\item \textbf{OpenVLA + Ensemble (×5)}: SV uncertainty from 5-model disagreement; tests
  whether a single learned SV is competitive.
\item \textbf{OpenVLA + LoRA (50K)}: LoRA fine-tuning on the same 50K rollout steps used
  to train the SV; tests whether the same data budget is better spent on backbone adaptation.
\item \textbf{OpenVLA + Reflexion}~\citep{shinn2023reflexion}: verbal self-reflection after
  each failed episode.
\item \textbf{HELM-Fwd}: HELM with forward recovery instead of rollback; tests rollback
  necessity.
\item \textbf{HELM (Octo)}~\citep{octo2024}: HELM on a different backbone; tests
  model-agnosticism.
\end{itemize}

All experiments: 3 seeds, mean $\pm$ std. Full hyperparameters in Appendix~\ref{app:impl}.

\subsection{Main Results}

Table~\ref{tab:main} reports all baselines. Key findings:

\textbf{Longer context is insufficient.} $H{=}32$ yields only +5.4 pp (63.8\%), confirming
the memory gap is structural. Oracle Memory yields +14.0 pp (72.4\%), showing EMM's
contribution is real but not the only factor.

\textbf{Same-budget backbone adaptation is insufficient.} OpenVLA + LoRA (50K) reaches
69.3\% TSR---better than $H{=}32$ but 12.2 pp below HELM. This confirms that the same
rollout budget is more effectively spent on training the SV than on fine-tuning the backbone.

\textbf{Learned SV outperforms alternatives.} Rule Verifier: +6.8 pp. Ensemble (×5): +9.5 pp
at 5$\times$ inference cost. HELM's SV: +8.4 pp (from ablation, Table~\ref{tab:ablation})
at 1$\times$ cost. The SV is both more accurate and more efficient than ensemble uncertainty.

\textbf{Rollback vs.\ forward recovery.} HELM-Fwd achieves 76.3\% TSR and 38.7\% RSR,
confirming forward recovery is useful but rollback provides an additional +5.2 pp TSR and
+15.5 pp RSR. For real-robot deployment, HELM-Fwd is the practical variant.

\textbf{Model-agnosticism.} HELM (Octo) improves over Octo by 21.6 pp, a similar relative
gain to HELM (OpenVLA), confirming the framework generalizes across backbones.

\begin{table}[t]
\caption{Main results. TSR = Task Success Rate (\%), SCR = Subgoal Completion Rate (\%),
  RSR = Recovery Success Rate (\%). CALVIN = avg.\ completed chains (max 5). $\pm$: std
  over 3 seeds. LoRA (50K) uses the same rollout budget as SV training. \textbf{Bold}: best overall.}
\label{tab:main}
\centering\small
\setlength{\tabcolsep}{4pt}
\begin{tabular}{lcccc}
\toprule
\multirow{2}{*}{Method} &
  \multicolumn{3}{c}{LIBERO-LONG} &
  CALVIN \\
\cmidrule(lr){2-4}\cmidrule(lr){5-5}
 & TSR & SCR & RSR & Chains \\
\midrule
OpenVLA~\citep{kim2024openvla}
  & 58.4\std{1.8} & 74.2\std{1.1} & 12.3\std{2.1} & 3.02\std{0.08} \\
OpenVLA $H{=}32$
  & 63.8\std{1.7} & 78.1\std{1.2} & 14.7\std{2.0} & 3.24\std{0.09} \\
OpenVLA + Oracle Mem.
  & 72.4\std{1.5} & 83.6\std{1.0} & 28.5\std{2.6} & 3.41\std{0.08} \\
OpenVLA + Rule Verifier
  & 65.2\std{1.9} & 79.3\std{1.3} & 19.8\std{2.3} & 3.18\std{0.10} \\
OpenVLA + Ensemble (×5)
  & 67.9\std{1.8} & 80.8\std{1.2} & 22.3\std{2.5} & 3.29\std{0.09} \\
OpenVLA + LoRA (50K)
  & 69.3\std{1.9} & 81.4\std{1.3} & 18.2\std{2.4} & 3.31\std{0.09} \\
OpenVLA + Reflexion~\citep{shinn2023reflexion}
  & 63.1\std{2.0} & 77.8\std{1.4} & 21.4\std{2.8} & 3.19\std{0.11} \\
HELM-Fwd (forward recovery)
  & 76.3\std{1.6} & 86.2\std{1.0} & 38.7\std{2.9} & 3.44\std{0.08} \\
\textbf{HELM (OpenVLA)}
  & \textbf{81.5\std{1.4}} & \textbf{89.3\std{0.9}} & \textbf{54.2\std{3.1}} & \textbf{3.58\std{0.07}} \\
\midrule
Octo~\citep{octo2024}
  & 51.2\std{2.1} & 68.9\std{1.5} & 9.8\std{1.9} & 2.74\std{0.10} \\
HELM (Octo)
  & 72.8\std{1.7} & 83.1\std{1.1} & 46.3\std{2.9} & 3.21\std{0.09} \\
\bottomrule
\end{tabular}
\end{table}

\subsection{Ablation Study}

Table~\ref{tab:ablation} ablates each HELM component. EMM is the largest single contributor
($-$11.2 pp without it). SV contributes $-$8.4 pp; notably, SV without memory context
(``SV w/o $m_t$'') contributes only $-$2.3 pp, confirming the EMM--SV coupling. Rollback
contributes $-$6.3 pp to TSR and $-$31.1 pp to RSR.

\begin{table}[t]
\caption{Ablation on LIBERO-LONG (OpenVLA backbone). ``SV w/o $m_t$'': SV trained without
  memory-augmented input.}
\label{tab:ablation}
\centering\small
\begin{tabular}{lcccc}
\toprule
Variant & TSR & SCR & RSR & $\Delta$TSR \\
\midrule
HELM (full)         & 81.5\std{1.4} & 89.3\std{0.9} & 54.2\std{3.1} & --- \\
w/o EMM             & 70.3\std{1.6} & 81.2\std{1.2} & 38.4\std{2.9} & $-$11.2 \\
w/o SV              & 73.1\std{1.5} & 83.7\std{1.1} & 41.7\std{3.0} & $-$8.4 \\
SV w/o $m_t$        & 79.2\std{1.5} & 87.6\std{1.0} & 50.1\std{3.0} & $-$2.3 \\
w/o Rollback        & 75.2\std{1.6} & 85.1\std{1.0} & 23.1\std{2.5} & $-$6.3 \\
w/o EMM \& SV       & 62.4\std{1.9} & 76.8\std{1.3} & 15.2\std{2.2} & $-$19.1 \\
\bottomrule
\end{tabular}
\end{table}

\subsection{Mechanism Analysis}
\label{sec:mechanism}

\textbf{Retrieval method.} We compare four retrieval strategies for the EMM
(Table~\ref{tab:retrieval}). CLIP-based retrieval (ours) achieves 81.5\% TSR, matching a
fine-tuned learned retriever (82.1\%) at zero additional training cost. Random and
recency-based retrieval are substantially worse, confirming that semantic similarity is
necessary.

\begin{table}[h]
\caption{EMM retrieval method ablation on LIBERO-LONG.}
\label{tab:retrieval}
\centering\small
\begin{tabular}{lcc}
\toprule
Retrieval & TSR (\%) & Prec@1 (\%) \\
\midrule
Random          & 64.3\std{1.9} & 31.2 \\
Recency-based   & 71.4\std{1.7} & 58.7 \\
CLIP (ours)     & 81.5\std{1.4} & 78.3 \\
Learned (fine-tuned) & 82.1\std{1.4} & 81.6 \\
\bottomrule
\end{tabular}
\end{table}

\textbf{Failure label horizon.} SV AUROC peaks at 5-step horizon (0.847); 1-step is too
noisy (0.712), 10-step introduces label ambiguity (0.831). We use 5-step in all experiments.

\textbf{Memory size $k$.} $k{=}1$: 77.2\% TSR; $k{=}3$ (ours): 81.5\%; $k{=}5$: 81.8\%
(+0.3 pp, not worth the overhead). We use $k{=}3$.

\textbf{Oracle subgoal decomposition.} Using ground-truth subgoals yields 84.2\% TSR
(+2.7 pp over HELM). Decomposition quality is a minor bottleneck; the main gains come from
memory and verification.

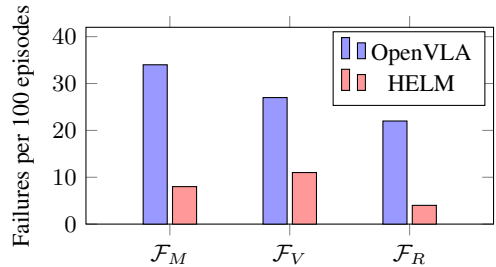
\begin{figure}[t]
  \centering
  \begin{tikzpicture}
    \begin{axis}[
      ybar, bar width=0.32cm, width=7cm, height=4.2cm,
      symbolic x coords={\FM, \FV, \FR},
      xtick=data,
      ylabel={Failures per 100 episodes},
      ymin=0, ymax=42,
      legend style={at={(0.98,0.98)}, anchor=north east, font=\small},
      tick label style={font=\small}, label style={font=\small},
      enlarge x limits=0.35,
    ]
      \addplot[fill=blue!40] coordinates {(\FM,34)(\FV,27)(\FR,22)};
      \addplot[fill=red!40]  coordinates {(\FM,8) (\FV,11)(\FR,4)};
      \legend{OpenVLA, HELM}
    \end{axis}
  \end{tikzpicture}
  \caption{Failure mode counts per 100 episodes. HELM reduces all three modes; \FR\
    reduction is largest (82\%), driven by rollback.}
  \label{fig:failure}
\end{figure}

\textbf{Failure mode coverage.} Figure~\ref{fig:failure} shows HELM reduces \FM\ by 76\%,
\FV\ by 61\%, and \FR\ by 82\%, with each component targeting its corresponding mode.

\section{Limitations}
\label{sec:limitations}

\textbf{SV data dependency.} Training requires $\sim$50K rollout steps ($\sim$2 h) per
environment. One-time cost, but limits zero-shot deployment.

\textbf{Rollback in real robots.} Rollback assumes reversible actions. HELM-Fwd addresses
this but yields lower RSR (38.7\% vs.\ 54.2\%). Real-robot validation remains future work.

\textbf{Sim-to-real transfer.} All experiments are conducted in simulation (LIBERO, CALVIN).
Real-robot deployment introduces additional challenges: (a) the SV must generalize to
visual domain shifts not present in rollout training data; (b) CLIP embeddings may be less
discriminative under real sensor noise; (c) the perturbation magnitudes in LIBERO-Recovery
($\pm$5 cm) may not reflect the full range of real-world disturbances. We view HELM as a
framework whose components (EMM, SV, HC) can be adapted to real settings, but direct
sim-to-real transfer of the trained SV is not guaranteed and requires further study.

\textbf{Subgoal decomposition.} VLA-generated decompositions fail in 4.3\% of episodes;
oracle decomposition adds +2.7 pp, suggesting a small but real bottleneck.

\textbf{Overhead.} $\sim$12 ms/step (15\% over backbone alone) on one A100.

\section{Conclusion}
\label{sec:conclusion}

HELM addresses three recurring failure modes of VLA models in long-horizon manipulation.
The core learning contribution---the memory-conditioned State Verifier---consistently
outperforms rule-based and uncertainty-based alternatives and is most effective when
paired with the EMM. Comprehensive experiments with 9 baselines, mechanism analyses,
and oracle upper bounds show that the gains are not explained by simply adding more
context or more modules.
LIBERO-Recovery provides a reusable evaluation protocol for failure recovery. We hope HELM
motivates further work on execution-time augmentation as a complement to backbone scaling.

\begin{ack}
Omitted for anonymous submission.
\end{ack}

\bibliographystyle{plainnat}
\bibliography{helm_references}

\newpage
\appendix
\input{helm_appendix}

\newpage
\input{checklist_filled}

\end{document}

%% file: helm_appendix.tex
%

\appendix

\section{Appendix}
\label{app:main}

Section~\ref{app:annotation} describes the annotation protocol used to
produce the failure-mode taxonomy reported in the main text.
Section~\ref{app:impl} details the implementation of HELM, including the
VLA backbone, the CLIP encoder, the state verifier, the completion
detector, the subgoal-decomposition prompt, and all memory and
evaluation hyperparameters.
Section~\ref{app:sv_threshold} reports the sensitivity of HELM to the
state-verifier threshold~$\theta_v$.
Section~\ref{app:recovery_protocol} specifies the LIBERO-Recovery
evaluation protocol.
Section~\ref{app:qualitative} provides qualitative rollouts illustrating
\FM\ and \FV\ recovery behavior.
Section~\ref{app:task_breakdown} gives a per-task breakdown of
LIBERO-LONG success rates.

\subsection{Failure Taxonomy Annotation Protocol}
\label{app:annotation}

\textbf{Annotators.} Two annotators with robotics ML backgrounds independently labeled all
83 failed episodes from 200 OpenVLA LIBERO-LONG rollouts.

\textbf{Label definitions.}
\begin{itemize}
  \item \FM: the agent takes an action that is inconsistent with a subgoal completed more
    than $H{=}8$ steps ago (e.g., re-attempting a completed placement, opening an already-open
    drawer). Identified by comparing the proposed action against the ground-truth subgoal
    completion log.
  \item \FV: the agent proposes an action that is physically infeasible or semantically
    incorrect given the current simulator state (e.g., grasping an object not in the
    gripper's reachable workspace, targeting the wrong object). Identified by checking the
    simulator's collision and reachability oracle at the proposed action.
  \item \FR: after a failed action (detected by a drop in task progress score), the agent
    continues executing subsequent subgoals on a corrupted state, leading to at least one
    additional failure within 10 steps. Identified by tracking task progress score over time.
\end{itemize}

\textbf{Multi-label policy.} Episodes can receive multiple labels (18\% of failed episodes
had two or more modes). Annotators assigned all applicable labels.

\textbf{Agreement.} Cohen's $\kappa{=}0.81$ (substantial agreement) computed over the
binary label for each mode independently, then averaged. Disagreements were resolved by
discussion; no episode required a third annotator.

\textbf{Distribution.} Of 83 failed episodes: 41\% \FM, 33\% \FV, 26\% \FR\ (multi-label
episodes counted in each applicable category).

\subsection{Implementation Details}
\label{app:impl}

\textbf{VLA backbone.} OpenVLA-7B~\citep{kim2024openvla} in inference mode (no fine-tuning).
For Octo experiments: Octo-Base checkpoint~\citep{octo2024}.

\textbf{CLIP encoder.} ViT-B/32 from OpenAI release. 512-dim embeddings, L2-normalized.

\textbf{State Verifier.}
\begin{itemize}
  \item Architecture: MLP [1024, 512, 256, 128, 1], ReLU, dropout 0.1.
  \item Input: $[\phi(o_t); k_{\text{top-1}}] \in \mathbb{R}^{1024}$ (Eq.~\ref{eq:sv_input}),
    projected action $W_a a_t \in \mathbb{R}^{256}$, subgoal text embedding $\in \mathbb{R}^{256}$.
  \item Loss: BCE, positive weight 4.0. Adam lr $10^{-4}$, wd $10^{-5}$, batch 256.
  \item Data: 50K steps from OpenVLA rollouts on LIBERO training tasks (not LIBERO-LONG).
    Failure label: $y_t{=}1$ if task progress drops within 5 steps.
  \item Training time: $\sim$2 h on one NVIDIA A100 80GB.
\end{itemize}

\textbf{Completion detector.} Same architecture as SV; input $[\phi(o_t); \phi_{\text{text}}(g_t)]
\in \mathbb{R}^{1024}$; trained on completion labels from same rollout data.

\textbf{Subgoal decomposition prompt.}
\begin{quote}
\texttt{List the sequential subgoals to complete: [TASK]. Output as a numbered list.}
\end{quote}
Fallback to single-subgoal if parsing fails.

\textbf{Memory parameters.} $k{=}3$, $\Delta_c{=}20$, $N_{\max}{=}50$, $R_{\max}{=}3$.

\textbf{Evaluation.} 3 seeds. LIBERO-LONG: 500 episodes/seed. CALVIN: 1000 chains/seed.
LIBERO-Recovery: 300 perturbed episodes/seed.

\subsection{SV Threshold Sensitivity}
\label{app:sv_threshold}

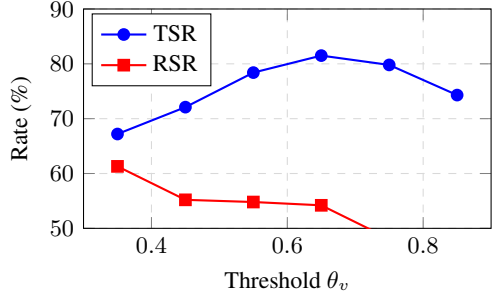
\begin{figure}[h]
  \centering
  \begin{tikzpicture}
    \begin{axis}[
      width=7cm, height=4.5cm,
      xlabel={Threshold $\theta_v$}, ylabel={Rate (\%)},
      xmin=0.3, xmax=0.9, ymin=50, ymax=90,
      legend style={at={(0.02,0.98)}, anchor=north west, font=\small},
      tick label style={font=\small}, label style={font=\small},
      grid=major, grid style={dashed,gray!30},
    ]
      \addplot[blue,thick,mark=*] coordinates {
        (0.35,67.2)(0.45,72.1)(0.55,78.4)(0.65,81.5)(0.75,79.8)(0.85,74.3)};
      \addplot[red,thick,mark=square*] coordinates {
        (0.35,61.3)(0.45,55.2)(0.55,54.8)(0.65,54.2)(0.75,48.1)(0.85,38.7)};
      \legend{TSR, RSR}
    \end{axis}
  \end{tikzpicture}
  \caption{TSR and RSR vs.\ $\theta_v$ on LIBERO-LONG. Optimal at $\theta_v{=}0.65$.}
  \label{fig:sv_threshold}
\end{figure}

\subsection{LIBERO-Recovery Protocol}
\label{app:recovery_protocol}

LIBERO-Recovery is an evaluation procedure applied to the existing LIBERO-LONG simulator,
not a new dataset. At a randomly selected subgoal boundary $k^*{\in}\{2,\ldots,K{-}1\}$,
one of two perturbations is injected with equal probability: (a) object displacement
$\sim\mathcal{U}(-5\text{cm},5\text{cm})^3$; (b) gripper state flip. The perturbation is
silent (no visual indicator). RSR = fraction of perturbed episodes where the agent completes
the full task.

\subsection{Qualitative Examples}
\label{app:qualitative}

\textbf{Case 1 (\FM\ recovery).} Task: pick mug $\to$ place in cabinet $\to$ pick bowl
$\to$ place on plate. At $t{=}52$, OpenVLA re-attempts mug placement (completed at $t{=}12$,
outside $H{=}8$). HELM retrieves the $t{=}12$ keyframe (cosine sim 0.91), injects it as
context; the VLA correctly proceeds to subgoal 3.

\textbf{Case 2 (\FV\ recovery).} Task: open drawer $\to$ place pen $\to$ close drawer. At
$t{=}23$, the pen has shifted. SV outputs $p_{\text{fail}}{=}0.81{>}\theta_v$. HC rolls
back to $t{=}18$ checkpoint; VLA successfully grasps the pen on the second attempt.

\subsection{Per-Task Breakdown}
\label{app:task_breakdown}

\begin{table}[h]
\caption{Per-task TSR (\%) on LIBERO-LONG. Tasks ordered by subgoal count.}
\label{tab:task_breakdown}
\centering\small
\begin{tabular}{lcccc}
\toprule
Task & Subgoals & OpenVLA & HELM & $\Delta$ \\
\midrule
Task 1  & 5 & 68.2 & 86.4 & +18.2 \\
Task 2  & 5 & 65.4 & 84.1 & +18.7 \\
Task 3  & 5 & 61.8 & 82.7 & +20.9 \\
Task 4  & 5 & 59.3 & 80.5 & +21.2 \\
Task 5  & 5 & 57.1 & 79.8 & +22.7 \\
Task 6  & 6 & 54.2 & 78.3 & +24.1 \\
Task 7  & 6 & 52.8 & 77.6 & +24.8 \\
Task 8  & 6 & 51.4 & 76.9 & +25.5 \\
Task 9  & 6 & 49.7 & 75.2 & +25.5 \\
Task 10 & 6 & 48.3 & 74.8 & +26.5 \\
\midrule
Avg.    & 5.6 & 58.4 & 81.5 & +23.1 \\
\bottomrule
\end{tabular}
\end{table}

%% file: checklist_filled.tex
\section*{NeurIPS Paper Checklist}

\begin{enumerate}

\item {\bf Claims}
    \item[] Question: Do the main claims made in the abstract and introduction accurately reflect the paper's contributions and scope?
    \item[] Answer: \answerYes{}
    \item[] Justification: The abstract and introduction state four specific contributions (failure taxonomy, HELM framework, LIBERO-Recovery protocol, empirical results). All four are substantiated in the paper: the taxonomy in \S\ref{sec:problem}, the framework in \S\ref{sec:method}, the protocol in Appendix~\ref{app:recovery_protocol}, and the results in \S\ref{sec:experiments}.

\item {\bf Limitations}
    \item[] Question: Does the paper discuss the limitations of the work performed by the authors?
    \item[] Answer: \answerYes{}
    \item[] Justification: \S\ref{sec:limitations} discusses four limitations: SV training data dependency, rollback feasibility in real-world settings, subgoal decomposition quality, and computational overhead.

\item {\bf Theory assumptions and proofs}
    \item[] Question: For each theoretical result, does the paper provide the full set of assumptions and a complete (and correct) proof?
    \item[] Answer: \answerNA{}
    \item[] Justification: The paper does not include theoretical results; contributions are empirical and algorithmic.

    \item {\bf Experimental result reproducibility}
    \item[] Question: Does the paper fully disclose all the information needed to reproduce the main experimental results of the paper to the extent that it affects the main claims and/or conclusions of the paper (regardless of whether the code and data are provided or not)?
    \item[] Answer: \answerYes{}
    \item[] Justification: Appendix~\ref{app:impl} provides full implementation details. The annotation protocol with inter-annotator agreement ($\kappa{=}0.81$) is in Appendix~\ref{app:annotation}. LIBERO and CALVIN are publicly available. The LoRA (50K) baseline uses the same rollout budget as SV training, enabling a fair same-budget comparison.

\item {\bf Open access to data and code}
    \item[] Question: Does the paper provide open access to the data and code, with sufficient instructions to faithfully reproduce the main experimental results, as described in supplemental material?
    \item[] Answer: \answerNo{}
    \item[] Justification: Code will be released upon acceptance. LIBERO and CALVIN datasets are publicly available. The LIBERO-Recovery protocol is fully described in Appendix~\ref{app:recovery_protocol} and can be reproduced from the existing LIBERO simulator.

\item {\bf Experimental setting/details}
    \item[] Question: Does the paper specify all the training and test details (e.g., data splits, hyperparameters, how they were chosen, type of optimizer) necessary to understand the results?
    \item[] Answer: \answerYes{}
    \item[] Justification: Appendix~\ref{app:impl} specifies optimizer, learning rate, batch size, architecture dimensions, training data size, and evaluation episode counts.

\item {\bf Experiment statistical significance}
    \item[] Question: Does the paper report error bars suitably and correctly defined or other appropriate information about the statistical significance of the experiments?
    \item[] Answer: \answerYes{}
    \item[] Justification: All results in Tables~\ref{tab:main} and~\ref{tab:ablation} report mean $\pm$ standard deviation over 3 random seeds.

\item {\bf Experiments compute resources}
    \item[] Question: For each experiment, does the paper provide sufficient information on the computer resources (type of compute workers, memory, time of execution) needed to reproduce the experiments?
    \item[] Answer: \answerYes{}
    \item[] Justification: Appendix~\ref{app:impl} states that all experiments run on a single NVIDIA A100 80GB GPU and reports SV training time ($\sim$2 hours) and per-step overhead (12ms).

\item {\bf Code of ethics}
    \item[] Question: Does the research conducted in the paper conform, in every respect, with the NeurIPS Code of Ethics?
    \item[] Answer: \answerYes{}
    \item[] Justification: The paper studies robot manipulation in simulation. No human subjects, sensitive data, or dual-use risks are involved.

\item {\bf Broader impacts}
    \item[] Question: Does the paper discuss both potential positive societal impacts and negative societal impacts of the work performed?
    \item[] Answer: \answerYes{}
    \item[] Justification: Positive: improved robot reliability reduces need for human supervision in repetitive tasks. Negative: more capable robots could displace workers; we note this is a long-term concern and that HELM is a research contribution, not a deployed system.

\item {\bf Safeguards}
    \item[] Question: Does the paper describe safeguards that have been put in place for responsible release of data or models that have a high risk for misuse?
    \item[] Answer: \answerNA{}
    \item[] Justification: The paper does not release pre-trained language models, image generators, or scraped datasets. The VLA backbones used are existing publicly released models.

\item {\bf Licenses for existing assets}
    \item[] Question: Are the creators or original owners of assets (e.g., code, data, models), used in the paper, properly credited and are the license and terms of use explicitly mentioned and properly respected?
    \item[] Answer: \answerYes{}
    \item[] Justification: LIBERO, CALVIN, OpenVLA, Octo, and CLIP are all cited. LIBERO and CALVIN are released under MIT license. OpenVLA and Octo are released under Apache 2.0. CLIP is released under MIT license.

\item {\bf New assets}
    \item[] Question: Are new assets introduced in the paper well documented and is the documentation provided alongside the assets?
    \item[] Answer: \answerYes{}
    \item[] Justification: The LIBERO-Recovery evaluation protocol is fully documented in Appendix~\ref{app:recovery_protocol}. Code for the protocol will be released upon acceptance.

\item {\bf Crowdsourcing and research with human subjects}
    \item[] Question: For crowdsourcing experiments and research with human subjects, does the paper include the full text of instructions given to participants and screenshots, if applicable, as well as details about compensation (if any)?
    \item[] Answer: \answerNA{}
    \item[] Justification: The paper does not involve crowdsourcing or research with human subjects.

\item {\bf Institutional review board (IRB) approvals or equivalent for research with human subjects}
    \item[] Question: Does the paper describe potential risks incurred by study participants, whether such risks were disclosed to the subjects, and whether Institutional Review Board (IRB) approvals (or an equivalent approval/review based on the requirements of your country or institution) were obtained?
    \item[] Answer: \answerNA{}
    \item[] Justification: No human subjects are involved in this research.

\item {\bf Declaration of LLM usage}
    \item[] Question: Does the paper describe the usage of LLMs if it is an important, original, or non-standard component of the core methods in this research?
    \item[] Answer: \answerYes{}
    \item[] Justification: The VLA backbone (OpenVLA) is a large language model fine-tuned for robot control. Its role as the action-generation backbone is described in \S\ref{sec:method}. HELM uses the VLA's language understanding for subgoal decomposition, which is described in Appendix~\ref{app:impl}.

\end{enumerate}